# Causal Reinforcement Learning for Optimisation of Robot Dynamics in Unknown Environments

Julian Gerald Dcruz, Sam Mahoney, Jia Yun Chua, Adoundeth Soukhabandith, John Mugabe, Weisi Guo, Miguel Arana-Catania
*School of Aerospace, Transport and Manufacturing*
Cranfield University
United Kingdom
{julian.geralddcruz.451, sam.mahoney.467, jiayun.chua.065, a.soukhabandith.289, john.mugabe.352, weisi.guo, miguel.arana-catania}@cranfield.ac.uk

*Abstract*—Autonomous operations of robots in unknown environments are challenging due to the lack of knowledge of the dynamics of the interactions, such as the objects' movability. This work introduces a novel Causal Reinforcement Learning approach to enhancing robotics operations and applies it to an urban search and rescue (SAR) scenario. Our proposed machine learning architecture enables robots to learn the causal relationships between the visual characteristics of the objects, such as texture and shape, and the objects' dynamics upon interaction, such as their movability, significantly improving their decision-making processes. We conducted causal discovery and RL experiments demonstrating the Causal RL's superior performance, showing a notable reduction in learning times by over 24.5% in complex situations, compared to non-causal models.

*Keywords—causal learning, reinforcement learning, autonomous systems, robotics*

## I. Introduction

Recently, the convergence of causality and reinforcement learning (RL) has become a vibrant area for progress, especially in crafting autonomous systems. Causality [1-3], the study of cause and effect, offers a more nuanced understanding of the environment than traditional correlation-based approaches. In RL, an agent learns to make decisions by interacting with its environment and receiving feedback in the form of rewards or penalties. Integrating causality into this framework allows agents to infer the causal relationships between their actions and the outcomes, leading to more robust decision-making processes. This causal understanding is crucial for tasks requiring long-term planning and for operating in dynamic or previously unseen environments. This project aims at enhancing efficiency in critical situations such as search and rescue (SAR) operations in smart cities using Causal Reinforcement Learning. To address this, we have developed a novel methodology that allows robots to learn about these causal relationships through direct interaction with their surroundings. By observing and manipulating objects, the robots can discern patterns and connections between what they see and how things move. This equips the robots with the foresight to anticipate outcomes and adapt swiftly, thereby accelerating the completion of their tasks and improving global multi-agent understanding of a collapsed world model during diverse disasters.

## II. Related Work

Yin et al. [4] aimed to improve the navigation of an autonomous agent without prior knowledge of the environment using off-policy RL. They used a Soft Actor-Critic with Curriculum Prioritization and Fuzzy Logic to assess and plan the navigation trajectory of the agent. To counteract the sim-to-real transfer problem, the authors proposed the use of Generative Adversarial Networks [5]. Similarly, regarding the use of artificial intelligence to assist SAR missions. Zuluaga et al. [6] have adapted an RL module for UAVs to improve the efficiency of SAR missions. However, most existing UAVs performing these tasks rely on greedy or potential-based heuristics without the ability to learn. These approaches are often inaccurate in real-world applications; furthermore, they require knowledge of the search space beforehand. The authors have also acknowledged the difficulties of using this agent in high-dimensional continuous state/action spaces where the RL agent finds it difficult to find an optimal policy. In this project, we aim to further reinforce this by applying causal learning [7-9] to the agent on top of an RL module to further improve its efficiency.

Sontakke et al. [10] proposed causal curiosity as "a novel intrinsic reward", capable of allowing an agent "to learn optimal sequences of actions and discover causal factors in the dynamics of the environment". They found that agents using causal learning can reduce the data required by the RL agents 2.5 times. The goal of an agent with causal curiosity is to allow agents to discover causal processes through exploratory interaction rather than focusing on maximizing the task reward. This project is focused on causal learning in enhancing robots' decisions [11] in SAR missions. Our causal learning work will apply the NOTEARS (Non-combinatorial Optimization via Trace Exponential and Augmented lagRangian for Structure learning) algorithm [12]. It transforms the challenge of learning a causal graph structure into a smooth, solvable optimization problem using gradient-based techniques. The benefits of this approach include scalability to larger datasets and a more efficient convergence towards accurate causality models. It assumes that the noise factors are Gaussian and that there are no hidden confounders. NOTEARS favours simpler, sparse graphs, employing regularization to reinforce this sparsity. Conventional parametric models presuppose predetermined mathematical relationships between the variables, while this nonparametric model is more able to conform to the underlying data structure.

Hu, T. [13] has applied simultaneous localization and mapping (SLAM), navigation algorithms, and a YOLOv7 neural network for object detection and depth estimation to create a robot capable of exploring unknown dynamic environments. SLAM can create a realistic map of the robot's surroundings and YOLOv7 is then used to detect humans in the environment; the robot then tries to predict the human's trajectory in the environment. An alternative way to localize the robot in its environment and map its surroundings is the long-term static mapping (LSM) and

This work was supported by EPSRC "TAS-S: Trustworthy Autonomous Systems: Security" (EP/V026763/1)

cloning localization (CL) method using a 3D LiDAR sensor [14]. LSM oversees the creation of an initial static 2D grid map and 3D feature map, whilst CL tracks and matches the 3D feature map to the dynamic changes in the environment.

## III. SYSTEM DESCRIPTION

### A. System Overview

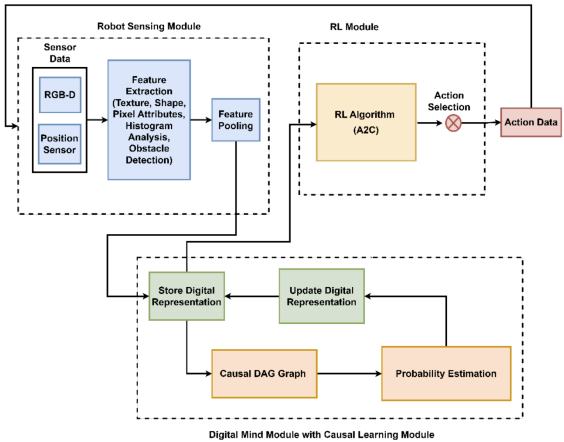

Fig. 1. Overarching system architecture proposed

Fig. 1 presents a depiction of the overarching system architecture proposed. It includes the Robot Sensing Module, which captures and processes environmental data, the RL Module, which guides the robot's action selection, and the Digital Mind Module, where causal learning and digital representation storage occur.

The Robot Sensing Module harnesses sensor data from two primary sources: RGB visual sensors and position sensors, providing an understanding of the robot's environment, and its orientation and movement. The collected data undergo feature extraction, where characteristics such as texture, shape, and pixel attributes are discerned, along with histogram analysis and the detection of obstacles. These data allow it to distinguish between different materials and surfaces and helps the robot make informed decisions about how to interact with different objects. Post-extraction, feature pooling is utilised to reduce dimensionality, preparing the data for more efficient processing. This ability is crucial in disaster scenarios where quick assessment and identification of survivors and paths through debris are essential. It should also be noted that the integration of the Sensing Module with the Digital Mind allows for a dynamic feedback loop where visual data inform the robot's causal learning processes and enables the robot to adapt its strategies based on visual clues and learned experiences.

The RL Module enables the robot to move and interact with the objects in the environment. As the robot is deployed in an unknown environment, it must autonomously understand and navigate through it. RL involves learning to make decisions by taking actions that maximise a cumulative reward. Here we employ an Advantage Actor-Critic (A2C) algorithm [15]. It makes use of two models: The actor takes actions in the environment based on its current policy and the critic assesses the action by calculating the value function of the resulting state generating a temporal difference error.

The Digital Mind allows the robot to perceive, understand, and interact with its surroundings meaningfully. It integrates sensor data to create a multidimensional representation of the surroundings, aiding navigation and interaction by identifying obstacles and spatial layouts. Central to this project is its ability for causal learning, where the robot engages with objects and observes outcomes to deduce physical laws. By pushing and moving objects, it learns causal relationships and predicts action consequences, forming Directed Acyclic Graphs (DAG) to model these relationships. The causal discovery [7-9] algorithm used is NOTEARS. The module estimates event probabilities based on experiences and observed causal relationships. Additionally, it interacts with the RL Module, updating digital world representations influenced by the robot's actions, allowing ongoing learning and refinement of causal models and decision-making processes.

A digital mind vector is initialised at the start of each episode, which stores the observations of the robot. The digital mind vector helps keep track of the object types the robot has interacted with, its previous actions, the movement status of the object if it has moved upon interaction, and the causal probability of the object. The digital mind vector is thus part of the robot's memory system. The movement status of the objects is tracked by checking if the object has moved upon interaction. Initially, with an unformed causal model, the robot is implemented to assume all objects have a 50% chance of being movable. This allows for updating each object type in the digital mind with their movement status. Each object texture type with its movement status is pulled from the digital mind and fed as input to the NOTEARS algorithm. The structural model created from the NOTEARS algorithm is then fed into the Bayesian model to predict the causal probability of the object being movable. The A2C RL algorithm with a Multi-Input Policy generates actions to be taken by the robot in the environment. The observation space information includes object positions, relative goal positions, collision data, and the history of previous actions. The robot's action space is restricted to forwards, backwards, left, and right. Inputs from computer vision, the digital mind, and causal probability contribute to constructing the robot's observation space.

## IV. EXPERIMENTS

To simulate a SAR mission, we implemented an environment with varying numbers of objects with different types of textures and shapes. A robot with a camera had to find its way to the trapped individual/goal by moving the blocked objects in its pathway or avoiding immovable objects in its pathway. In the experiments, we test independently the two main modules of our architecture. The code of the experiments is publicly available[1].

### A. Causal Discovery

This experiment is designed to assess the effectiveness of the NOTEARS causal discovery algorithm in accurately inferring from data the causal relationships between the texture and shape of the objects and their movability. It evaluates the number of samples (here interactions with the objects) required for the algorithm to consistently infer the true causal graph.

---
[1]https://github.com/Causal-Curiosity-in-Search-and-Rescue/Causal_RL_for_Robotics_in_Unknown_Environments

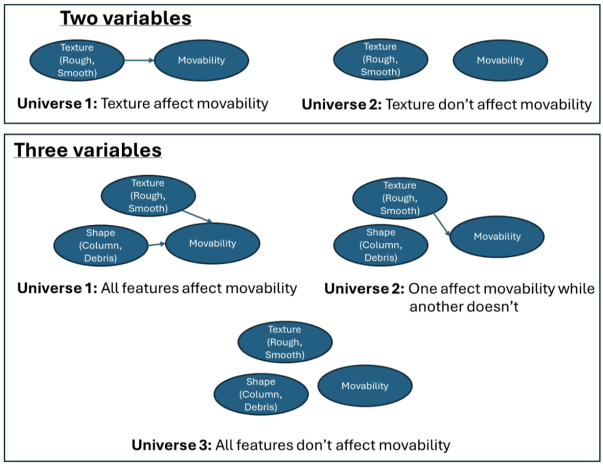

Fig. 2.  Causal relationships for all scenarios analysed

Fig. 2 illustrates the diverse universes under investigation, encompassing scenarios with two and three variables where texture or shape are causally related or not to movability. For each of these scenarios, synthetic datasets are generated with varying numbers of samples to mimic different quantities of observational data. Data are randomly produced across two categories for each of ten distinct datasets per data point.

For each dataset size, the NOTEARS algorithm is applied to infer the causal graph. This is performed 10 times for 10 different datasets to introduce variability and robustness in the results. The Structural Hamming Distance (SHD) and precision metrics are calculated for each inferred graph compared to the true graph. The SHD quantifies the number of discrepancies between the inferred and true graphs, with 0 indicating a perfect match. Precision measures the proportion of correctly inferred edges out of all edges inferred by the algorithm, with 1 indicating all inferences are correct. The implementation is done using CausalNex[2], NetworkX[3], and CausalDiscoveryToolbox[4] [16].

### B. Causal Discovery Results

This section showcases the SHD and precision over the number of samples under the various universes.

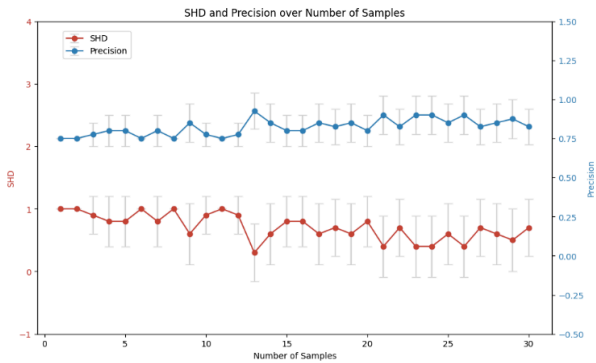

Fig. 3.  SHD/precision vs samples (2 causally related variables)

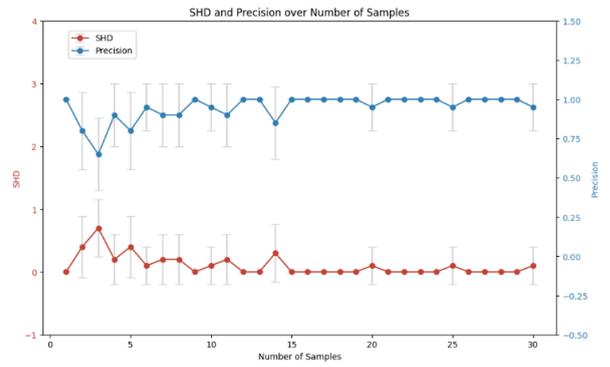

Fig. 4.  SHD/precision vs samples (2 independent variables)

Figs. 3 and 4 show the causal discovery efficacy over the number of samples in the two variables scenario where they are causally linked or disconnected.

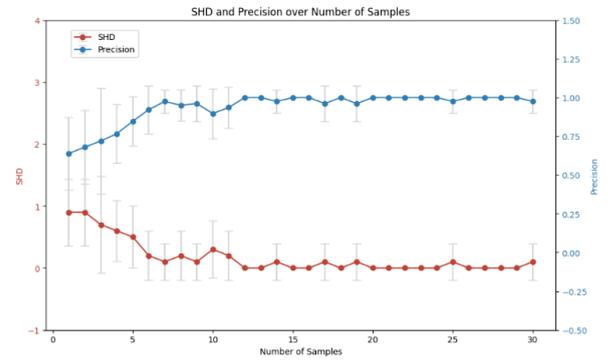

Fig. 5.  SHD/precision vs samples (3 variables with 2 causally related variables)

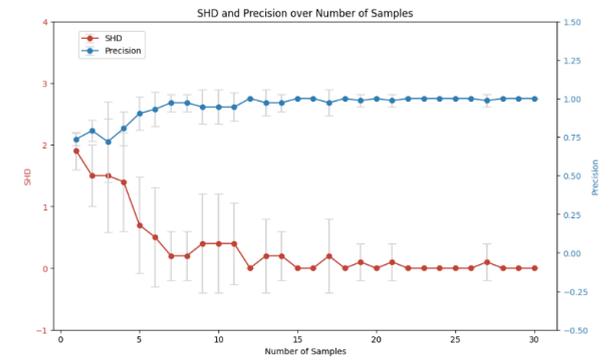

Fig. 6.  SHD/precision vs samples (3 causally related variables)

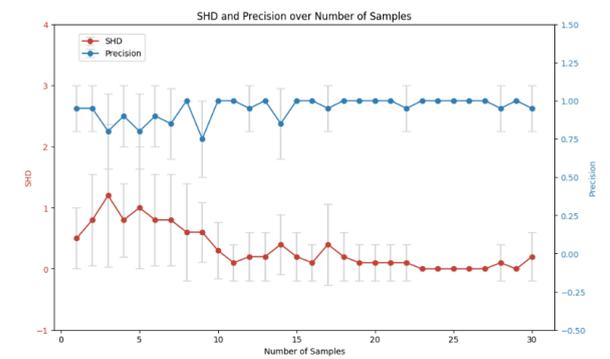

Fig. 7.  SHD/precision vs samples (3 independent variables)

---

[2] https://github.com/mckinsey/causalnex
[3] https://github.com/networkx
[4] https://github.com/FenTechSolutions/CausalDiscoveryToolbox

Figs. 5, 6, and 7 correspond to the three variables scenarios, with 2, 1, and 0 variables causally linked to each other. The results with three variables exhibit greater variability, stemming from the increased number of possible permutations, compared to the two variables case.

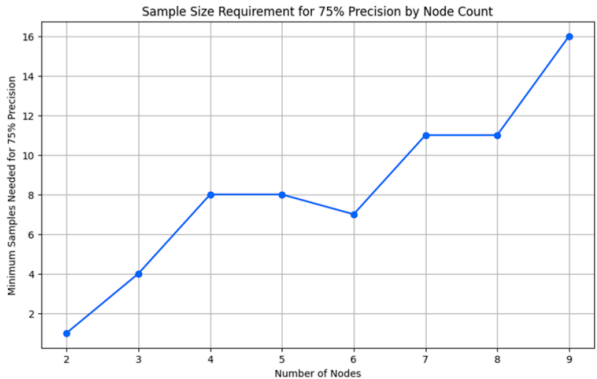

Fig. 8. Sample size requirement for 75% precision.

To explore whether an increase in variables impacts the minimum number of samples needed, we selected a graph in which one variable (such as texture) is causally linked to movability. We then incrementally added new independent variables (increasing the node count) and recorded the minimum number of samples required to achieve 75% precision. Fig. 8 shows the minimum samples needed for 75% precision as the number of variables increases.

### C. Causal Discovery Discussion

The previous results indicate that, for two variables, fewer samples are typically required to reach high precision. In contrast, the graph involving three variables shows increased variability due to a greater number of potential permutations. This requires a larger sample size to ensure stable convergence with high precision/low SHD.

Fig. 8 illustrates that to reach an initial precision of 0.75, only one sample is needed for two variables, while three variables require four samples. Additionally, beyond initial precision, the analysis also delved into stability, examining how consistently the results converge. The data suggest that an approximate quantity of 13.5 observations, as seen in Figs. 3 and 4, is required to accurately infer the stable relational structure among two variables, achieving 0.3 SHD and 0.9 precision. When examining three variables, the number of necessary observations, as seen in Figs. 5, 6 and 7, incrementally rises to 16. This parameter will be a fundamental assumption in the subsequent RL experiments.

Graph structures comprising a limited number of variables exhibit fewer permutational outcomes, resulting in diminished variability within the potential graph configurations. In contrast, an increment in the variable count grows exponentially the potential edge permutations and thus the inferential complexity. The graphs reveal that the standard deviation is significantly higher in the graphs with three variables compared to those with two variables. Scenarios with a large number of variables demand a proportionately larger sample size to efficiently discern the true causal model, as delineated in Fig. 8.

In certain universes, it appears that a system with three variables might yield more precise results eventually than one with just two. This improvement could stem from the fact that the additional variable provides more data and contextual relationships, which NOTEARS can utilize to better identify the true graph structure.

### D. Causal Reinforcement Learning

In the causal reinforcement learning experiments, we evaluate how causal relations formed by the agent in a highly complex ever-changing environment, as is in SAR, help provide on-policy RL algorithms with extra information in the observation space to solve the task more efficiently. In these experiments, we compare an agent that learns the causal relationships of the environment (using causal discovery procedures such as the one shown in the previous experiments) with a non-causal agent that does not have this knowledge. In a real-world SAR example, this causal knowledge may assist a robot for example in identifying which pieces of rubble are movable, which are hazardous, and which doors open and in what way.

Using the Gymnasium library we created a random 2D environment where the trapped individual (goal) is enclosed in a room with no straightforward openings to enter the room. Movable objects to simulate blocked openings or doors are placed as room boundaries, so the agent is required to move the objects located on the boundaries of the room to enter, as shown in Fig. 9.

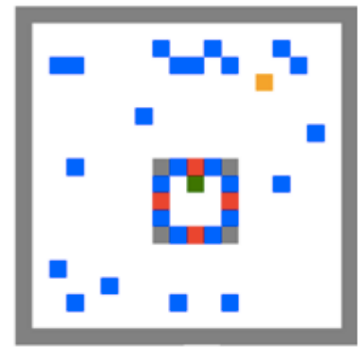

Fig. 9. Example of simulation environment showing immovable/movable (blue/red) objects, and robot/goal (yellow/green).

In these experiments we evaluate the scenarios described in Fig. 2, same as in the previous section, with the exception of the two trivial scenarios where none of the variables are connected to the movability.

In the implementation of the environment, factors such as the positions of the starting agent and the objects are random, to help the agents generalise their learning.

The reward function used is the following:

$$reward = min\left(10 + \left(\frac{800 - currentstep}{800}\right) \times 10, 20\right) \quad (1)$$

A (20,20) grid size was used for the environment with a (7,7) inner room located in the middle of the environment with the goal inside of it. The agent actions are {forward, backwards, left, right} 1 cell movements. Each episode has 800 maximum number of steps, which represent an out-of-fuel scenario or out-of-range scenario in a SAR mission. The observation space is of a Box type implemented as an integer ranging from 0 to 8; the corresponding objects are the following: wall, free space, rough debris, rough column, smooth debris, smooth column, agent starting position, and goal.

The causal agent observation space includes information about each object's movability, since we assume here that the previous causal discovery phase was fully successful for this agent, while for the non-causal agent it is not included.

The experiments are evaluated using the following aggregated metrics across the parallel environments: Mean number of times the goal has been reached (MGR), Mean time taken to reach the goal (MTT), Mean number of movable object interactions (MMI), Mean number of non-movable object interactions (MII).

The RL algorithm used is A2C with a Multilayer Perceptron Policy. The hyperparameters used are discount factor gamma=0.995, number of steps to run each environment per update n_steps=100, entropy coefficient ent_coef=0.002, value function coefficient vf_coef=0.5, maximum gradient norm max_grad_norm=1, learning rate lr=0.0003, and Adam optimizer with epsilon value=1e-7. The training is conducted over 8 million timesteps, with an evaluation interval at every 10,000 timesteps and a logging interval at every 5,000 timesteps, in an Nvidia Tesla T4 GPU with 16GB VRAM, 64GiB CPU RAM with 8 cores.

*E. Causal Reinforcement Learning Results*

In the first experiment there are 2 variables (texture, movability) and 2 types of objects: Rough objects which are immovable and smooth objects which are movable. The number of objects is varied in the simulations. The results are shown in Table I and Fig. 10. All plots in this section have applied a smoothing of 10% with a running average method to understand the general trend for comparison, and correspond to 18 objects.

TABLE I. Causal RL - 2 variables causally connected

| Agent | No. of objects | MGR | MTT | MMI | MII |
|---|---|---|---|---|---|
| Causal | 6 | **0.98** | 0.52 | 4.44 | 8.70 |
| Non-Causal | 6 | 0.93 | **0.31** | **3.27** | **5.87** |
| Causal | 12 | **1.00** | 0.29 | **2.74** | **7.72** |
| Non-Causal | 12 | 0.95 | 0.29 | 3.43 | 8.31 |
| Causal | 18 | **1.00** | **0.20** | **2.97** | **6.55** |
| Non-Causal | 18 | 0.93 | 0.28 | 3.08 | 10.24 |

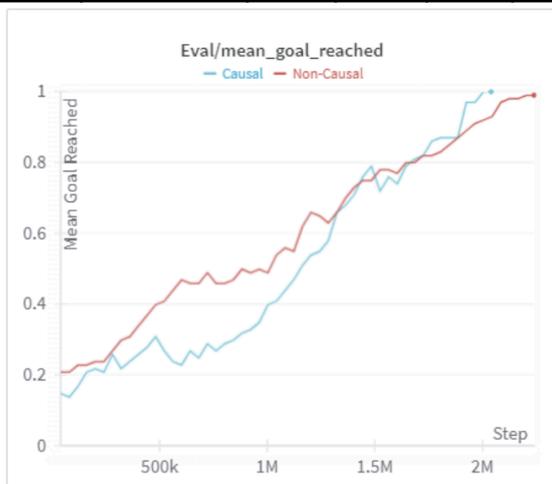

Fig. 10. MGR for 2 variables causally connected scenario

Fig. 10 shows both models learning, however, the causal models consistently reach the early stopping of mean_goal_reached=1 earlier than the non-causal models. As seen in Table I, in the more complex environments with 18 objects, the causal agent showed a better performance across all metrics conforming with our hypothesis that the additional causal knowledge on object movability helps the agent navigate the environment more efficiently.

In the second experiment, we used 3 variables (texture, shape, movability) and 4 types of objects, either rough or smooth textured, and shaped like debris or columns. In this case, the texture of the objects is causally connected to movability and the shape is disconnected. The results are shown in Table II and Fig. 11.

TABLE II. Causal RL - 3 variables partially connected

| Agent | No. of objects | MGR | MTT | MMI | MII |
|---|---|---|---|---|---|
| Causal | 6 | **1.00** | 0.38 | **3.42** | **5.50** |
| Non-Causal | 6 | 0.94 | **0.29** | 3.48 | 5.53 |
| Causal | 12 | **1.00** | **0.21** | 3.53 | **7.77** |
| Non-Causal | 12 | 0.99 | 0.29 | **2.97** | 8.74 |
| Causal | 18 | **0.98** | **0.29** | 2.92 | **8.50** |
| Non-Causal | 18 | 0.76 | 0.58 | **2.61** | 19.41 |

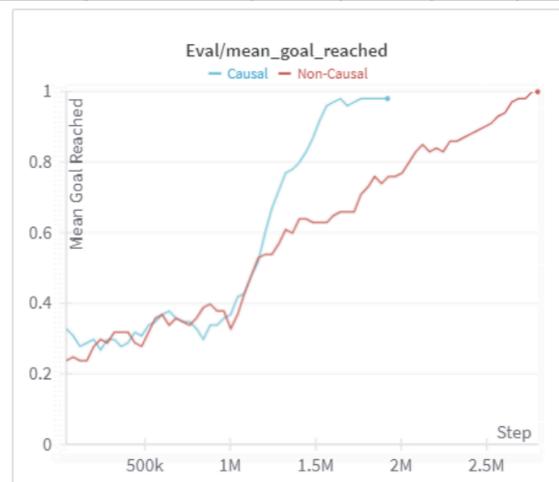

Fig. 11. MGR for 3 variables only partially connected scenario

Table II shows that for causal agents where the environment has objects with texture and shape, the more objects there are in the environment, the wider the difference in mean time taken and mean immovable interactions, and the causal agent has a significantly higher mean number of times goal has been reached compared to the non-causal agent. Fig. 11 shows how, with 18 objects, the causal agent shows a significant trend in better learning to reach the goal with optimized interactions with the movable and immovable objects in the environment.

In the third experiment, the same previous 3 variables (texture, shape, movability) and 4 types of objects are used. Here both the texture type and the shape are causally connected to movability and the inner room position is randomised each epoch creating a much more complex environment. Results are shown in Table III and Fig. 12.

TABLE III. CAUSAL RL - 3 VARIABLES CAUSALLY CONNECTED AND RANDOM GOAL POSITION

| Agent | No. of objects | MGR | MTT | MMI | MII |
|---|---|---|---|---|---|
| Causal | 18 | **0.90** | 0.73 | **4.16** | **49.67** |
| Non-Causal | 18 | 0.72 | **0.68** | 10.69 | 51.91 |

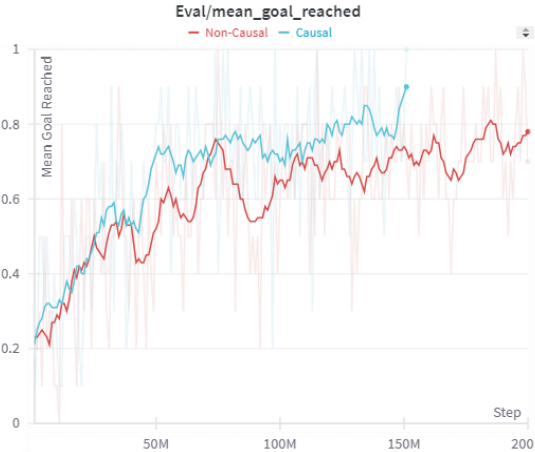

Fig. 12. MGR for 3 variables causally connected and random goal position scenario

Fig. 12 shows the causal model reached the early stopping criteria at around 150M time-steps, whilst the non-causal model exceeded the 200M limit without reaching it. The causal model was able to train at a minimum of 24.5% faster than the non-causal model.

In a second evaluation of these models, the causal and non-causal models created at 151M time-steps are tested on 9 random environments. The causal model successfully reached the goal in these tests 70% of the time, with an average completion time of 3 seconds, whilst the non-causal model reached the goal in these tests 60% of the time, with an average completion time of 9 seconds.

*F. Causal Reinforcement Learning Discussion*

The results have demonstrated improved learning times across all tested environments, which included using fixed goal and random goal positions and differing numbers of objects and their classifying variables. It is observed across all the runs that as the environment becomes more complex with an increasing number of objects, the causal agent outperforms the non-causal agent with a significant gap across most of the metrics. Unexpectedly, the causal model did not show a large reduction in the number of interactions, but this will be addressed in future work via further exploration of additional reward strategies. We theorise that one of the major limitations in our model comes from the restricted field of view of the robot agent, which allows the agent to observe only the space directly in front of it.

The model can be improved further by better tuning the RL hyperparameters for our task and through the use of penalties applied when interacting with the immovable objects in the environment. We believe that applying this penalty may encourage the causal model to interact less with the immovable objects with no difference in movable interactions, whilst in the non-causal model, we may see a decrease in all interactions.

## V. CONCLUSION

In this work we have proposed a new architecture for an autonomous robot combining sensing capabilities with causal understanding and autonomous decision-making via RL to enhance its efficiency. The system's main proposed goal is its use in SAR scenarios.

To validate our proposal, experiments were conducted on causal discovery and RL to explore the influence of causal knowledge on robot performance. We explored the system's ability to understand causal relationships, interpret visual and physical attributes of objects—like texture and shape—and predict the dynamic behaviour of unknown objects.

The causal discovery experiments using NOTEARS yielded key findings regarding the number of interactions necessary to learn the causal relationships of the environment, setting the stage for the subsequent RL simulations. Using the A2C algorithm we observed a remarkable improvement in learning times, with the Causal RL model accelerating task completion by over 24.5% in complex scenarios. While this research opens the door to future enhancements, the insights gleaned thus far underscore the significant promise of integrating causal learning with RL in SAR contexts.